\documentclass[conference]{IEEEtran}
\IEEEoverridecommandlockouts
\usepackage{cite}
\usepackage{amsmath,amssymb,amsfonts}
\usepackage{algorithmic}
\usepackage{graphicx}
\usepackage{textcomp}
\usepackage{xcolor}
\usepackage{booktabs}
\usepackage{tikz}
\usepackage{pgfplots,pgfplotstable}
\usetikzlibrary{patterns.meta}
\usetikzlibrary{positioning,fit,arrows.meta,backgrounds}
\usetikzlibrary{backgrounds}
\usetikzlibrary{patterns}
\def\BibTeX{{\rm B\kern-.05em{\sc i\kern-.025em b}\kern-.08em
    T\kern-.1667em\lower.7ex\hbox{E}\kern-.125emX}}

\definecolor{blue0}{RGB}{0,0,150}
\definecolor{blue1}{RGB}{30,60,255}
\definecolor{blue2}{RGB}{60,120,255}
\definecolor{blue3}{RGB}{0,184,230}
\definecolor{blue4}{RGB}{194,194,255}
\definecolor{blue5}{RGB}{38,77,115}

\begin{document}

\title{Analyzing the Traffic of MANETs using Graph Neural Networks\\
}

\author{\IEEEauthorblockN{Taha Tekdogan}
\IEEEauthorblockA{\textit{Department of Computer Engineering} \\
\textit{Istanbul Technical University}\\
Istanbul, Turkey \\
tekdogan20@itu.edu.tr}
}

\maketitle

\begin{abstract}
Graph Neural Networks (GNNs) have been taking role in many areas, thanks to their expressive power on graph-structured data.
On the other hand, Mobile Ad-Hoc Networks (MANETs) are gaining attention as network technologies have been taken to the 5G level.
However, there is no study that evaluates the efficiency of GNNs on MANETs.
In this study, we aim to fill this absence by implementing a MANET dataset in a popular GNN framework, i.e., PyTorch Geometric; and show how GNNs can be utilized to analyze the traffic of MANETs.
We operate an edge prediction task on the dataset with GraphSAGE (SAG) model, where SAG model tries to predict whether there is a link between two nodes.
We construe several evaluation metrics to measure the performance and efficiency of GNNs on MANETs.
SAG model showed 82.1\% accuracy on average in the experiments.
\end{abstract}


\section{Introduction}
Recently, representing data in non-euclidean domain is highly preferred.
Graphs are the most suitable data structures for expressing the topology among objects.
Therefore, learning from graph representations are gaining attention in many areas.
Graph Neural Networks (GNNs) are defacto methods for learning from graphs.

There have been many implementations of GNNs on a myriad of disciplines, from quantum chemistry\cite{Gilmer2017} to computational neuroscience\cite{Li2020}.
Thanks to advanced techniques of graph processing on Graphical Processing Units (GPUs)\cite{Che:pannotia:2013,Khorasani:CuSha:2014,Xu:2014:Graph,Wang:Gunrock:2016,Hong:MultiGraph:2017}, these tasks can be operated more efficiently in relatively much shorter times.
The success of such implementations encouraged the idea of applying GNN techniques on network technologies\cite{Boutaba2018,Rusek2019}.
There also have been a worldwide competition to solve network-related problems by using GNNs\cite{Suarez2021}.

One of the most popular wireless network specie is Mobile Ad-hoc NETwork (MANET), where nodes are dynamically connected to each other without any predefined routing protocol.
As a consequence of this behavior, each node is free to move in any direction, and may dynamically change its links between other nodes.
There are several techniques to investigate a MANET such as using a testbed or using software-based simulators.

Although the above-mentioned techniques tackle different problems of network-related challenges, to the best of our knowledge, there is no graph representation learning study that focuses on MANETs.
Therefore we decided to implement a representative MANET dataset (Stanford Gnutella dataset\cite{snap-dataset}) on a popular GNN framework (PyTorch Geometric\cite{pyg}) by using a highly expressive GNN model, i.e., GraphSAGE (SAG)\cite{sag}.
Then we conducted a research by running GNN pipelines on this dataset, and analyzed the traffic of this dataset by operating a link-prediction task.

We also construed a number of evaluation metrics to measure the performance and efficiency of used GNN model on implemented MANET dataset, i.e., execution time, accuracy, and scalability.
We deliver the results in Section~\ref{section:eval} and comment on the outcomes.

Our contributions are as follows:
\begin{itemize}
    \item We make the first GNN research on MANETs by delivering quantitative results of our experiments.
    \item We measure the performance of GNNs on MANET dataset by using our predefined evaluation metrics.
    \item We bring a new asset to the graph machine learning community by implementing a MANET dataset on a popular open-source GNN framework, i.e. PyTorch Geometric. Therefore it can be easily used for further studies without additional effort.
\end{itemize}

The remaining of this paper is structured as follows: Section II introduces the GNNs by delivering history and common notation, and then some fundamental concepts about MANETs are explained.
In Section III, we deliver our methodology by explaining our implemented dataset and used GNN model; and then the experimental setup is briefly described.
Section IV incorporates the evaluation of our experiments, i.e., methods for measuring the performance of GNNs on the MANET dataset, and results of the experiments by using these predefined evaluation metrics.
Finally, we conclude our study by commenting on the results and giving insightful suggestions inferred from our experiments.

\section{Background}

As the amount of data is excessing every day, there emerges a need to express it in various ways.
One type of such data is graphs, which are able to express the topology among objects and object's feature information.
Graphs are widely used in many areas to represent unstructured data such as molecules\cite{Gilmer2017}, social relationships\cite{Nettleton2013}, academic citations\cite{Jeong2020}, etc.

A graph $G$ consists of vertices $V$ and edges $E$ connecting vertices to each other.
Vertices in a graph may be connected to each other either directly or indirectly via edges.
Two vertices are neighbors if they are directly connected to each other.
The neighbor vertices of vertex $v$ is represented as $N(v)$.

Each vertex may carry information that we call it vertex attribute, and represent as $h_v$.
On the other hand, edges may carry information too; it is called edge attribute and represented as $g_e$.

GNN pipelines consist of one or more layers.
These layers are represented as $k~\epsilon~[1,L]$.
In each layer, the vertex and edge attributes are updated analogous to the given GNN formula.
For example, the vertex attribute of the layer $k$ is represented as ${h_v}^k$.

We provide all the above-mentioned notations we use during the study in Table~\ref{tab:notation}.

\begin{table}[]
    \centering
    \caption{Notations for graph neural networks}
    \begin{tabular}{cl}
        \toprule
        \textbf{Notation} & \textbf{Description} \\
        \midrule
        $G(V,E)$    & A graph \\
        $V$         & Set of vertices of the graph \\
        $E$         & Set of edges of the graph \\
        $v$         & A single vertex where $v~\epsilon~V$ \\
        $e$         & A single edge where $e~\epsilon~E$ \\
        $k$         & Present GNN layer where $k~\epsilon~[1,L]$ \\
        $h^{k}_v$ & Feature representation of node $v$ at layer $k$ \\
        $g^{k}_e$ & Feature representation of edge $e$ at layer $k$ \\
        $N(v)$      & Neighbourhood nodes of the node $v$ \\
        \bottomrule
    \end{tabular}
    \label{tab:notation}
\end{table}

\subsection{Graph Neural Networks}

GNNs are a type of neural network models where they can process and learn from graph structured data.
The term graph neural network (GNN) first coined by Scarselli et al.\cite{Scarselli2009} in 2009.
They propose a model to derive a supervised algorithm for estimating the parameters of GNN model.

There have been numerous GNN models proposed in the last decade, such as Graph Convolutional Network (GCN)\cite{gcn}, Graph Isomorphism Network (GIN)\cite{gin}, GraphSAGE (SAG)\cite{sag}, and Graph Attention Network (GAT)\cite{gat}.
Each GNN model has its characteristic computation complying with their mathematical formula.

\begin{figure}[h]
  \centering

\begin{tikzpicture}[
    show background rectangle]

    \node[circle, draw, fill=lightgray, font=\small]
    (I1) {$v_1$};
    \node[circle, below right=8mm and 1mm of I1, draw, fill=teal, font=\small]
    (I2) {$v_2$};
    \node[circle, below left=3mm and 5mm of I1, draw, fill=lime, font=\small]
    (I3) {$v_3$};
    \node[circle, above right=3mm and 3mm of I1, draw, fill=pink, font=\small]
    (I4) {$v_4$};
    \node[circle, above left=3mm and 3mm of I1, draw, fill=cyan, font=\small]
    (I5) {$v_5$};
    
    \node[rectangle, above=10mm of I1, font=\small]
    (T1) {input layer $1$};

    \draw[-, line width=0.3mm] (I2)--(I1);
    \draw[-, line width=0.3mm] (I3)--(I1);
    \draw[-, line width=0.3mm] (I4)--(I1);
    \draw[-, line width=0.3mm] (I3)--(I5);
    \draw[-, line width=0.3mm] (I4)--(I5);
    \draw[-, line width=0.3mm] (I2)--(I4);
    
%
%
    %




\end{tikzpicture}
\begin{tikzpicture}
    \node at (1,7) [circle,fill=white,inner sep=1.5pt](N1){};
    \node [above=15mm of N1, circle,fill,inner sep=1.5pt](N2){};
    \node [left=2mm of N2, circle,fill,inner sep=1.5pt](N3){};
    \node [right=2mm of N2, circle,fill,inner sep=1.5pt](N4){};
    \node [left=4mm of N2, circle,fill=white,inner sep=1.5pt](N5){};
    \node [right=4mm of N2, circle,fill=white,inner sep=1.5pt](N6){};
\end{tikzpicture}
\begin{tikzpicture}[
    show background rectangle]

    \node[circle, draw, fill=lightgray, font=\small]
    (I1) {$v_1$};
    \node[circle, below right=8mm and 1mm of I1, draw, fill=teal, font=\small]
    (I2) {$v_2$};
    \node[circle, below left=3mm and 5mm of I1, draw, fill=lime, font=\small]
    (I3) {$v_3$};
    \node[circle, above right=3mm and 3mm of I1, draw, fill=pink, font=\small]
    (I4) {$v_4$};
    \node[circle, above left=3mm and 3mm of I1, draw, fill=cyan, font=\small]
    (I5) {$v_5$};
    
    \node[rectangle, above=10mm of I1, font=\small]
    (T1) {output layer $N$};

    \draw[-, line width=0.3mm] (I2)--(I1);
    \draw[-, line width=0.3mm] (I3)--(I1);
    \draw[-, line width=0.3mm] (I4)--(I1);
    \draw[-, line width=0.3mm] (I3)--(I5);
    \draw[-, line width=0.3mm] (I4)--(I5);
    \draw[-, line width=0.3mm] (I2)--(I4);
    
%
%
    %




\end{tikzpicture}

\caption{An example schema of a GNN pipeline with $N$ layers.}
\label{fig-gnn}
\end{figure}
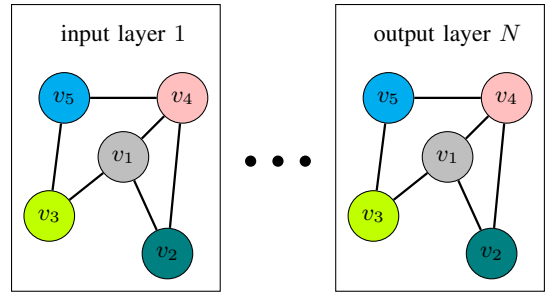

GNNs generally incorporates two main stages: inference and training.
Inference refers to aggregating neighbourhood information and updating each vertex attribute analogous to corresponding GNN model.
Training refers to updating model parameters and weights.
An illustration of the computation of a GNN pipeline with $N$ consecutive layers is given at Fig.~\ref{fig-gnn}.

Real-world graph datasets are getting huge in terms of size with millions of vertices, edges, and feature lengths.
This growth makes the computational cost of GNNs so expensive.
Due to their intensive, repeating, and parallellizable computation pattern, GNNs are generally conducted on GPUs.
There have been many studies for making graph processing and GNN computation efficient on GPUs\cite{Che:pannotia:2013,Khorasani:CuSha:2014,Xu:2014:Graph,Wang:Gunrock:2016,Hong:MultiGraph:2017}.

\subsection{MANETs}

\begin{figure}[htbp]
\centering
    \includegraphics[scale=1]{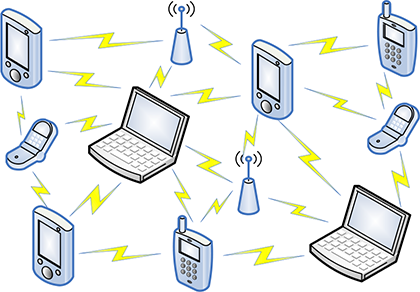}
\caption{An example of a MANET including different type of devices\cite{manet-website}.}
\label{fig-manets}
\end{figure}

Mobile Ad hoc NETworks (MANETs) are dynamic and decentralized networks colonized by mobile devices/stations.
These stations are generally devices such as mobile phones and laptops.
MANETs have a dynamic and self-configuring nature which changes continuously.
MANETs consist of many types of networks, such as Vehicle Ad hoc NETworks (VANETs) which are used in the communication of vehicles; Smartphone ad hoc networks (SPANs) which use existing infrastructure of Bluetooth and Wi-Fi connections; Wireless mesh networks are named after the topology of mesh networks; Army tactical MANETs are used for military purposes; Wireless sensor networks where sensors are nodes that create and convey information; and so on.
En example of a MANET is given at Fig.~\ref{fig-manets}\cite{manet-website}.

\section{Methodology}
This section introduces our methodology, i.e., dataset we used during the experiments, GNN model we used for learning, and the details about experimental setup.

\subsection{Dataset}
We use Stanford's Gnutella MANET dataset, which is Gnutella's peer-to-peer file sharing traffic from August 2002\cite{snap-dataset}.
We implemented the dataset in one of the most popular GNN frameworks, i.e., PyTorch Geometric\cite{pyg}.
Dataset consists of 8846 nodes and 31839 undirected edges.
Metadata information of the MANET dataset is given in Table~\ref{tab:dataset-info}.

\begin{table}
  \centering
  \caption{Metadata of the MANET Dataset}
  \label{tab:dataset-info}
  \begin{tabular}{cccc}
    \toprule
    \textbf{Dataset}     & \textbf{Nodes}  & \textbf{Feature Length}    & \textbf{Edges}  \\
    \midrule
    \textbf{MANET}\cite{snap-dataset}          & 8,846     & 1 & 31839\\
  \bottomrule
\end{tabular}
\end{table}

We operated link prediction task on our MANET dataset.
The task is to predict whether two nodes are connected to each other directly via an edge.
While implementing the dataset in the framework, we defined the train-test split mask as 80\% and 20\%, respectively.

\subsection{GNN Model}
We use the modified version of GraphSAGE (SAG)\cite{sag} model during the experiments.
SAG is a GNN model designed to inductively learn from current nodes of the graph, and predicts the unseen node features.
Even though SAG is designed to operate for node-prediction tasks, we made a modification on the model for working on link-prediction task\cite{online-link-pred}.

The formulation of SAG model for node classification is given at \eqref{sag-mp}.

\begin{equation}
\label{sag-mp}
    \centering\large {h_v^{(k)}} = W_1h_v^{(k-1)} + W_2\ast mean_{j\epsilon N(v)\cup\{v\}}h_u\normalsize
\end{equation}

In Eq.~\eqref{sag-mp}, $W_1$ is the weight of node $v$, and $W_2$ is the weight of neighbourhood nodes $N(v)$.
We exploit this formulation by giving a pair of nodes to model for link-prediction purpose.
Model takes two vertex attributes and calculates the probability of having an edge between these vertices, as given in \eqref{eq-prob}.

\begin{equation}
\label{eq-prob}
    \centering P(e_{vu}) = MLP(h_v \cdot h_u)
\end{equation}

We use the loss function given in \eqref{eq-loss} to maximize the probability of correct predictions during the training of our model.

\begin{equation}
\label{eq-loss}
    \centering loss = -mean(log(PP + \epsilon)) + mean(log(1 - NP + \epsilon))
\end{equation}

In Eq.~\eqref{eq-loss}, $PP$ represents positive predictions, and $NP$ means negative predictions.

\subsection{Experimental Setup}
Experiments were run ten times, and the average results of these runs were collected.
Experiments were conducted on Intel Xeon 2000 CPU and NVIDIA V100 GPU (32GB). 

We have four NVIDIA V100 GPUs in our environment, and they are linked to each other with an NVLink.
The number of GPUs participated during the experiments were configured dynamically due to \emph{scalability} metric (see Section~\ref{scalability}).

\section{Evaluation}
\label{section:eval}
In this section, we first define our evaluation metrics to measure the performance and success of GNN model on MANET dataset.
Then for each metric, we deliver quantitative results by commenting on outcomes.

\subsection{Evaluation Metrics}
\subsubsection{Accuracy}
Accuracy is the fraction of correctly predicted data over all predicted data.
We use this metric to show how successful the trained GNN model is on our MANET dataset.

\subsubsection{Scalability}
\label{scalability}
In real-world GNN tasks, as computation requirements are becoming intensive, many GPUs are used in parallel manner to improve the performance.
Adding more GPUs to the environment is decent as long as the performance is enhanced.
We use the metric scalability to examine this behaviour in our experiments.
We used $n$ GPUs where $n~\epsilon~\{1,2,4\}$, and observed the time taken to execute given GNN pipelines.

\subsubsection{Execution Time}
This metrics stands for the wall clock time taken for a GNN pipeline to accomplish its task.
We separately measure the duration of inference and training phases of GNN pipelines to show how they react to changes in environment (e.g. scalability).

\subsection{Results}
\begin{figure}[htbp]
\centering
    \includegraphics[scale=0.52]{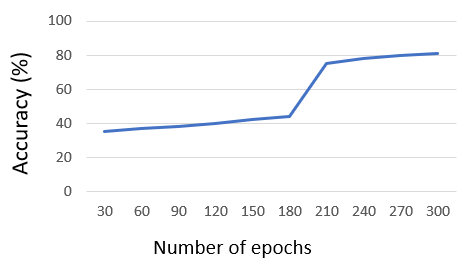}
\caption{Accuracy of the trained model over 300 epochs.}
\label{fig-epochs}
\end{figure}

During the experiments, we observed that our trained SAG model achieved 81.2\% accuracy on average, after 300 epochs.
The change over epochs can be observed in Fig.~\ref{fig-epochs}.

Furthermore, our experiments show that scaling the computing power by adding more GPUs to the environment does not always improve the performance.
Using four GPUs resulted in better performance ($\sim$38ms faster) than using one GPU; but showed worse performance ($\sim$22ms slower) than using two GPUs.
This outcome suggests that finding optimal value for the number of GPUs makes the pipeline efficient.

\begin{figure}[htbp]
\centering
    \includegraphics[scale=0.47]{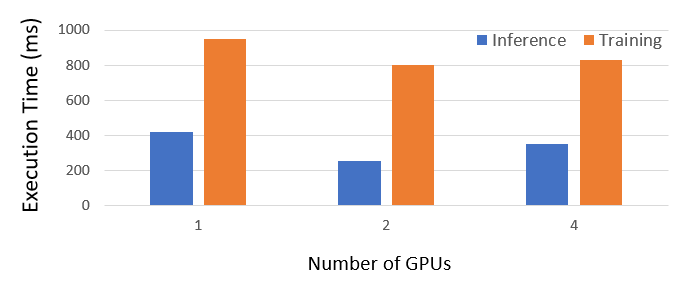}
\caption{Execution time of the GNN tasks with different GPU configurations.}
\label{fig-scalability}
\end{figure}

\section{Conclusion and Future Work}

We present an implementation of a representative MANET dataset on a popular GNN framework, i.e., PyTorch Geometric.
We construe a few evaluation metrics to measure the efficiency of GNNs on our implemented dataset.
Experiments results show that modified SAG model shows a success on analyzing the traffic of MANET dataset, which is a link prediction task, with 81.2\% accuracy on average.
Our results also suggest that increasing the number of GPUs is not always efficient as the overhead of distributing and accumulating operations overrides the bound of batch data size.

As a future work, we plan to operate node classification and graph classification tasks with proper network datasets.


\begin{thebibliography}{00}

\bibitem{Gilmer2017} J. Gilmer, S. S. Schoenholz, P. F. Riley, O. Vinyals, and G. E. Dahl, “Neural Message Passing for Quantum Chemistry,” arXiv [cs.LG], 2017.

\bibitem{Li2020} X. Li et al., “Pooling Regularized graph neural network for fMRI biomarker analysis,” Med. Image Comput. Comput. Assist. Interv., vol. 12267, pp. 625–635, 2020.

\bibitem{Che:pannotia:2013} S. Che, B. M. Beckmann, S. K. Reinhardt and K. Skadron, ``Pannotia: Understanding irregular GPGPU graph applications,'' 2013 IEEE International Symposium on Workload Characterization (IISWC), 2013, pp. 185-195, doi: 10.1109/IISWC.2013.6704684.

\bibitem{Khorasani:CuSha:2014} Farzad Khorasani, Keval Vora, Rajiv Gupta, and Laxmi N. Bhuyan, ``CuSha: vertex-centric graph processing on GPUs,'' In Proceedings of the 23rd international symposium on High-performance parallel and distributed computing (HPDC '14). Association for Computing Machinery, New York, NY, USA, 239–252. https://doi.org/10.1145/2600212.2600227

\bibitem{Xu:2014:Graph} Q. Xu, H. Jeon and M. Annavaram, ``Graph processing on GPUs: Where are the bottlenecks?,'' 2014 IEEE International Symposium on Workload Characterization (IISWC), 2014, pp. 140-149, doi: 10.1109/IISWC.2014.6983053.

\bibitem{Wang:Gunrock:2016} Yangzihao Wang, Andrew Davidson, Yuechao Pan, Yuduo Wu, Andy Riffel, and John D. Owens,``Gunrock: a high-performance graph processing library on the GPU,'' SIGPLAN Not. 51, 8, Article 11 (August 2016), 12 pages. https://doi.org/10.1145/3016078.2851145

\bibitem{Hong:MultiGraph:2017} C. Hong, A. Sukumaran-Rajam, J. Kim and P. Sadayappan, ``MultiGraph: Efficient Graph Processing on GPUs,'' 2017 26th International Conference on Parallel Architectures and Compilation Techniques (PACT), 2017, pp. 27-40, doi: 10.1109/PACT.2017.48.

\bibitem{Rusek2019} K. Rusek, J. Suárez-Varela, A. Mestres, P. Barlet-Ros, and A. Cabellos-Aparicio, “Unveiling the Potential of Graph Neural Networks for Network Modeling and Optimization in SDN,” in Proceedings of the 2019 ACM Symposium on SDN Research, 2019, pp. 140–151. doi: 10.1145/3314148.3314357.

\bibitem{Boutaba2018} R. Boutaba et al., “A comprehensive survey on machine learning for networking: evolution, applications and research opportunities,” Journal of Internet Services and Applications, vol. 9, no. 1, p. 16, Jun. 2018, doi: 10.1186/s13174-018-0087-2.

\bibitem{Suarez2021} J. Suárez-Varela et al., “The Graph Neural Networking Challenge: A Worldwide Competition for Education in AI/ML for Networks,” SIGCOMM Comput. Commun. Rev., vol. 51, no. 3, pp. 9–16, Jul. 2021, doi: 10.1145/3477482.3477485.

\bibitem{snap-dataset} J. Leskovec and A. Krevl, ``SNAP Datasets: Stanford Large Network Dataset Collection.'' http://snap.stanford.edu/data, 2014. [Online]. Available: http://snap.stanford.edu/data

\bibitem{pyg} M. Fey and J. E. Lenssen, “Fast Graph Representation Learning with PyTorch Geometric,” in ICLR 2019 Workshop on Representation Learning on Graphs and Manifolds, New Orleans, USA, 2019.

\bibitem{Nettleton2013} D. F. Nettleton, “Data mining of social networks represented as graphs,” Computer Science Review, vol. 7, pp. 1–34, 2013, doi: https://doi.org/10.1016/j.cosrev.2012.12.001.

\bibitem{Jeong2020} C. Jeong, S. Jang, E. Park, and S. Choi, “A context-aware citation recommendation model with BERT and graph convolutional networks,” Scientometrics, vol. 124, no. 3, pp. 1907–1922, 2020.

\bibitem{Scarselli2009} F. Scarselli, M. Gori, A. C. Tsoi, M. Hagenbuchner, and G. Monfardini, “The graph neural network model,” IEEE Trans. Neural Netw., vol. 20, no. 1, pp. 61–80, 2009.

\bibitem{gcn} Kipf, T. N. and Welling, M., ``Semi-Supervised Classification with Graph Convolutional Networks'', 2016.

\bibitem{gin} K. Xu, W. Hu, J. Leskovec, and S. Jegelka, “How Powerful are Graph Neural Networks?,” arXiv [cs.LG], 2018.

\bibitem{sag} W. L. Hamilton, R. Ying, and J. Leskovec, “Inductive representation learning on large graphs,” arXiv [cs.SI], 2017.

\bibitem{gat} Petar Veličković, Guillem Cucurull, Arantxa Casanova, Adriana Romero, Pietro Liò and Yoshua Bengio, ``Graph Attention Networks,'' 6th International Conference on Learning Representations, ICLR 2018, Vancouver, BC, Canada, April 30 - May 3, 2018. doi: 10.48550/ARXIV.1710.10903.

\bibitem{online-link-pred} T. Jain, “Online link prediction with Graph Neural Networks,” Medium, 16-Mar-2022. [Online]. Available: https://medium.com/stanford-cs224w/online-link-prediction-with-graph-neural-networks-46c1054f2aa4. [Accessed: 30-Aug-2022]. 


\bibitem{manet-website} ``Manet network in internet of things system,'' IntechOpen. [Online]. Available: https://www.intechopen.com/chapters/53178. [Accessed: 27-Aug-2022] DOI:10.5772/62746.


\end{thebibliography}
\end{document}